\documentclass[final]{cvpr}
\usepackage{multirow}
\usepackage{times}
\usepackage{epsfig}
\usepackage{graphicx}
\usepackage{amsmath}
\usepackage{amssymb}
\usepackage{indentfirst}
\usepackage{float}
\usepackage{flushend}
\usepackage{multirow}
\usepackage{xcolor}
\usepackage{colortbl}

\usepackage[pagebackref,breaklinks,colorlinks]{hyperref}

\def\cvprPaperID{3615} 
\def\confName{CVPR}
\def\confYear{2023}
\graphicspath{{./figs/}{./figs/result/}}

\begin{document}
\title{HumanDiffusion: a Coarse-to-Fine Alignment Diffusion\\ Framework for Controllable Text-Driven Person Image Generation}
\author{
Kaiduo Zhang\textsuperscript{1,2}, Muyi Sun\textsuperscript{1}, Jianxin Sun\textsuperscript{1,2}, Binghao Zhao\textsuperscript{1,2}, Kunbo Zhang\textsuperscript{1\footnotemark[2]}, Zhenan Sun\textsuperscript{1,2}, Tieniu Tan\textsuperscript{1,2,3}
\\
\textsuperscript{1} Center for Research on Intelligent Perception and Computing, NLPR, CASIA\\
\textsuperscript{2} School of AI, University of Chinese Academy of Sciences (UCAS) 
\textsuperscript{3} Nanjing University (NJU)\\
\tt\small \{kaiduo.zhang, muyi.sun, jianxin.sun, binghao.zhao\}@cripac.ia.ac.cn, \\ \tt\small \ kunbo.zhang@ia.ac.cn, \{znsun, tnt\}@nlpr.ia.ac.cn
}

\maketitle

\begin{abstract}
\vspace{-2mm}
Text-driven person image generation is an emerging and challenging task in cross-modality image generation. 
Controllable person image generation promotes a wide range of applications such as digital human interaction and virtual try-on. 
However, previous methods mostly employ single-modality information as the prior condition (e.g. pose-guided person image generation) or utilize the preset words for text-driven human synthesis. 
Introducing a sentence composed of free words with an editable semantic pose map to describe person appearance is a more user-friendly way. 
In this paper, we propose \textbf{HumanDiffusion}, a coarse-to-fine alignment diffusion framework, for text-driven person image generation.
Specifically, two collaborative modules are proposed, the Stylized Memory Retrieval (SMR) module for fine-grained feature distillation in data processing and the Multi-scale Cross-modality Alignment (MCA) module for coarse-to-fine feature alignment in diffusion.
These two modules guarantee the alignment quality of the text and image, from image-level to feature-level, from low-resolution to high-resolution. 
As a result, HumanDiffusion realizes open-vocabulary person image generation with desired semantic poses.
Extensive experiments conducted on DeepFashion demonstrate the superiority of our method compared with previous approaches.
Moreover, better results could be obtained for complicated person images with various details and uncommon poses. 
\end{abstract}
\vspace{-8mm}
\section{Introduction}
\vspace{-1mm}

High-fidelity and high-diversity person image generation is an emerging and challenging task in multi-modal interaction systems.
Previous approaches\cite{humangan,ren2022neural,cheong2022pose,zhu2019progressive,zhang2021keypoint,siarohin2018deformable} mostly focus on single-modality guided generation (e.g. skeleton-based or semantic-guided). 
Recently, with the rapid development of the cross-modality image generation\cite{anyface,ho2022classifier,gu2022vector,gafni2022make,lafite,diffusionclip,ramesh2022hierarchical}, some researchers try to explore the text-guided human image synthesis with the preset vocabularies \cite{jiang2022text2human,verbal-perosn,pami}.
However, their methods are limited in practical applications. More \textbf{flexible} and \textbf{free-style} text-driven person image generation is desirable to be explored.

\begin{figure}[t]
\centering
\includegraphics[width=0.48\textwidth]{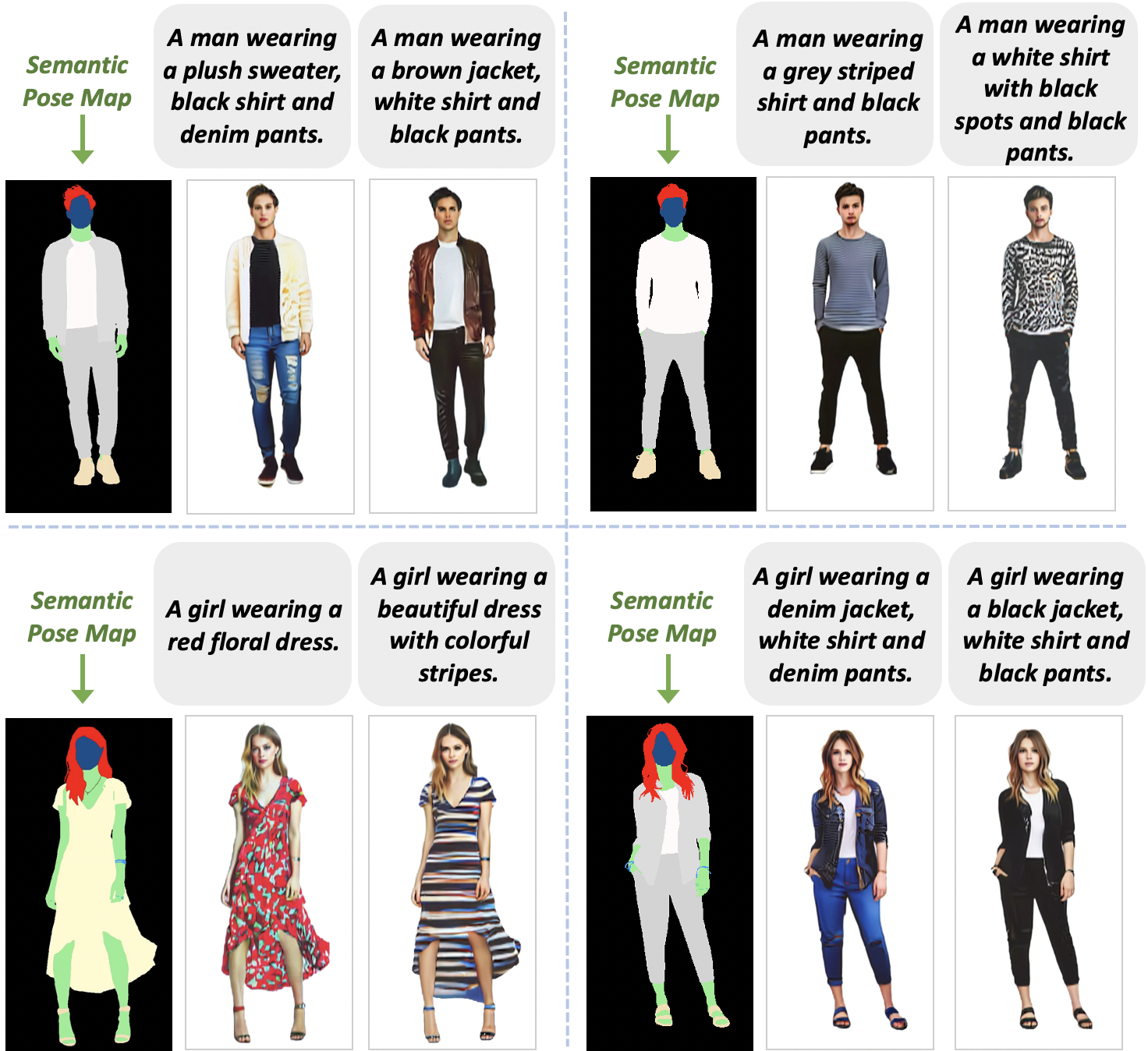}
\caption{The open-vocabulary driven person images generation task in this paper. Given the semantic pose maps and the appearance descriptions, HumanDiffusion can synthesize high-quality and high-fidelity person images in free-style.}
\label{Figure1}
\vspace{-8mm}
\end{figure}

In general, human synthesis methods are mostly implemented with traditional generation frameworks, such as Generative Adversarial Networks (GANs)\cite{humangan,he2022style,lewis2021tryongan,fu2022stylegan} or Variational Auto-Encoders (VAEs)\cite{jiang2022text2human}.
From the perspective of the target tasks, the human 
synthesis can be divided into two categories: human pose transfer \cite{ren2022neural,zhu2019progressive,zhang2021keypoint,zhang2021pise} and text-driven person image generation\cite{pami,jiang2022text2human,verbal-perosn}.
Pose transfer aims to transfer the human appearances from the source image into the desired pose, and commonly follows the GANs-based architecture. 
The main issue of human pose transfer is to align the cross-domain image features\cite{zhang2021pise}.
Though the generation quality of these methods is satisfactory, they are not intuitive and user-friendly in real scenes.
In contrast, as a carrier that can express information more flexibly, the text shows more interactivity and controllability for person image generation.
The rich expressions in the text bring more diversity to the human body in colors, textures, and even clothing types as shown in Figure \ref{Figure1}.







Recent text-driven person image generation methods have realized basic human image generation from input prompts which consisted of preset words\cite{verbal-perosn, pami, jiang2022text2human}. 
Verbal-Person\cite{verbal-perosn}, a GANs-based network, introduces the local text-relevant and global full-body discriminators to promote the text correspondence with the target skeleton under the supervision of the paired text annotations. 
Similarly, Image-Text Shared Space (ITSS)\cite{pami} proposes a flexible and integrated GANs-based framework for text-guided human image manipulation, which explores different moving distances in the shared latent space of the text and image. 
Different from the low-quality generation methods above mentioned \cite{verbal-perosn,pami}, Text2Human\cite{jiang2022text2human} designs a multi-stage framework for high-resolution text-driven human synthesis which utilizes a texture-aware codebook to realize fine-grained feature representation. 
However, these methods mostly use preset words as the supervision and show limitations in processing multi-modality feature fusion and alignment, which are not flexible for complex human image description and generation. Therefore, there is an urgent demand to introduce an open-vocabulary method for free-style human appearance description and appropriate frameworks for multi-modality fusion.

Recently, diffusion models such as denoising diffusion probabilistic models (DDPM)\cite{ho2020denoising} and score-based generative models\cite{song2019generative,song2020score} show powerful capability in image generation tasks\cite{nichol2021glide, gafni2022make, ramesh2022hierarchical}. 
The latest researches \cite{dhariwal2021diffusion, ramesh2022hierarchical, kawar2022imagic} perform better potential in text information perception and high-quality image generation, compared to VAEs\cite{doersch2016tutorial} and GANs\cite{goodfellow2020generative}. 
These days, diffusion models have attracted wide attention and have been applied to various fields, especially in cross-modality generation tasks.

For flexible and free-style person image generation, we propose \textbf{HumanDiffusion}, a newly designed diffusion-based framework with coarse-to-fine cross-modality feature alignment modules. 
In this coarse-to-fine framework, two collaborative modules are designed for high-quality image generation, a Stylized Memory Retrieval (SMR) module and a Multi-scale Cross-modality Alignment (MCA) module. 
The SMR module is located before the input appearance features, which introduces a disentangled memory network to refresh the semantic representations from low-quality information into fine-grained features and enhance the text features with the image-aware representations for sampling. Thus, the SMR module endows HumanDiffusion with the ability of text-free training and text-driven sampling.
The MCA module is designed with the cascaded cross-attention mechanism and integrated into the diffusion backbone, which gradually aligns the  middle-layer image-aware features and input text-aware features from low-resolution to high-resolution.
Furthermore, the HumanDiffusion is created with a compact structure and implemented with an end-to-end text-free training strategy.


The contributions are summarized as follows:
\begin{itemize}
    \item A coarse-to-fine alignment diffusion framework, named \textbf{HumanDiffusion}, is proposed for flexible, free-style, and text-driven person image generation. 
    \item A Multi-scale Cross-modality Alignment (MCA) module is designed into the diffusion framework, which gradually improves the feature alignment degrees of the text and image, from low-resolution to high-resolution. 
    \item A Stylized Memory Retrieval (SMR) module is proposed to provide more fine-grained image-aware partial features, which also enhances the open-vocabulary capability of text guidance. 
    \item Extensive experiments conducted on the DeepFashion dataset demonstrate the superiority of the HumanDiffusion. Better results are obtained for complicated person generation and manipulation.
\end{itemize}

\section{Related Works}

\begin{figure*}[htbp]
\includegraphics[width=1\textwidth]{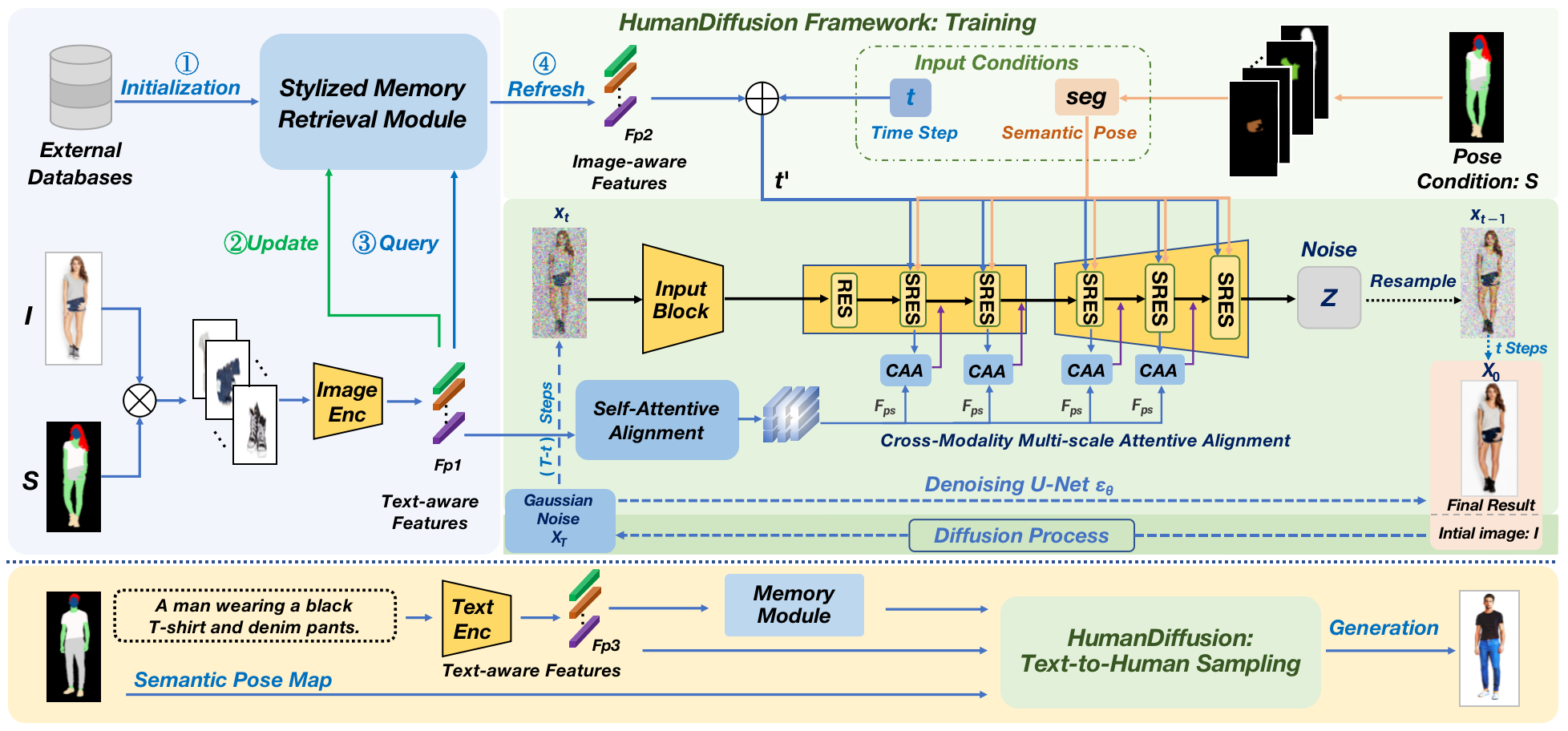} 
\caption{Overview of the HumanDiffusion framework. 
In the training procedure, the Stylized Memory Retrieval (SMR) module is designed to obtain fine-grained image-aware features which support text-free training; a diffusion-based network is proposed for cross-modality feature alignment with the help of the Multi-scale Cross-modality Alignment (MCA) module under the pose condition. 
In the sampling procedure, given a text description and a semantic pose map, a text-driven person image could be generated with HumanDiffusion. }
\vspace{-4mm}
\label{Figure2}
\end{figure*}
\subsection{Controllable Person Image Generation}

Controllable person image generation is a challenging vision task\cite{humangan,he2022style,fu2022stylegan,jiang2022text2human,lewis2021tryongan,ma2018disentangled,li2019dense,wang2022self}. 
Researchers try to apply different generative models to handle this problem.
Most existing models synthesize human images in two aspects, pose generation\cite{zhang2022exploring,zhu2019progressive,zhang2021keypoint,zhang2021pise} and texture refinement\cite{lewis2021tryongan,he2022style}. 
SAWN\cite{zhang2021controllable} uses a spatially-adaptive warped normalization to align the style and pose features. 
Pose-with-Style\cite{albahar2021pose} proposes a pose-conditioned StyleGAN\cite{fu2022stylegan} framework and wraps the target pose with the source image details. 
TryOnGAN\cite{lewis2021tryongan} proposes a semantic-driven model to synthesize human images for virtual try-on. 
However, these methods are not flexible, and can not be applied interactively in real scenes.
Text2Human\cite{jiang2022text2human} introduces a text-driven perspective for human synthesis. 
The Verbal-Person\cite{verbal-perosn} employs a structure-enhanced GAN model to generate similar text-guided human images. 
However, these GANs-based methods 
are limited in cross-modality feature alignment, and commonly need paired text annotations for training. 
To overcome these problems, we propose an end-to-end and text-free training framework to generate person images, conditioned on the target semantic poses and natural language descriptions.

\subsection{Text to Image Generation with Diffusion}

Recently, text-to-image generation tasks have made great progress with the emergence of Diffusion models\cite{nichol2021glide,gafni2022make,ramesh2022hierarchical,gu2022vector,liu2021more,ramesh2021zero}. Due to the ability to preserve the semantic structure of data, diffusion models can perform higher diversity and fidelity than previous generation frameworks\cite{diffusionclip}.
Many researches have explored text-conditional image generation, in which the most outstanding schemes should be the huge cross-modality models like DALLE-2\cite{ramesh2022hierarchical}, GLIDE\cite{nichol2021glide}. 
These models can generate amazing images according to the prompts people desire, which can even synthesize scenes that people have never seen. 
Furthermore, Retrieval-Diffusion\cite{retrieval} introduces the nearest neighbor retrieval to improve the generation quality and diversity. 
Imagic\cite{kawar2022imagic} performs better results for diverse text-driven image editing. 
However, these models are all general text-to-image frameworks on natural image generation which can not perceive fine-grained features for specific downstream tasks such as person image generation.
In this paper, we focus on the sub-task of text-driven person image generation and design a specific diffusion framework, which could deeply explore the human image properties (\emph{e.g.} semantic structural information).

\subsection{Memory Networks}
The idea of memory networks is mainly inherited from RNN, including LSTM\cite{hochreiter1997long} and GRU\cite{cho2014learning}. 
The memory networks have been applied to various vision tasks, including image captioning \cite{chunseong2017attend}, text-to-image synthesis\cite{zhu2019dm}. 
Weston et al.\cite{miao2020memory} propose memory networks to overcome the limitations of RNNs. 
Zhu et al.\cite{zhu2019dm} utilize a multi-modal memory module to refresh and clear blurred images for text-to-image generation. 
Recently, retrieval modules appear in text-to-image diffusion models such as KNN-Diffusion\cite{knn} and Retrieval-Diffusion\cite{retrieval}, which retrieve the nearest neighbors of the condition images.
These neighbors provide extra details to enhance the image expression.
However, they are mostly fixed and preset in the retrieval module, which limits the representative ability of fine-grained features.
Inspired by the memory mechanism, we design a dynamic memory module to provide text-aware features which can be more consistent with the distribution of the training data, and supply better additional detailed information which enhances clothes representation in free-style.
\section{Method}
HumanDiffusion is implemented on the diffusion-based U-Net framework as shown in Figure \ref{Figure2}, which consists of two parts: the Stylized Memory Retrieval (SMR) module and the Multi-scale Cross-modality Alignment (MCA) module. 
In the following, we will introduce this method in detail with the training and sampling procedure specifically. 
As for the training process, first given a human image $I$ and the corresponding semantic map $S$, the coarse features ${F_{p1}}$ are obtained by the CLIP image encoder.
The ${F_{p1}}$ can be also treated as the text-aware representations, which is attributed to the powerful alignment ability of CLIP.  
Thus, text-free training could be achieved.
Then, $F_{p1}$ is sent into SMR to update the memory blocks and retrieve the fine-grained appearance features ${F_{p2}}$, which is image-aware and combined with $t$ to get the global condition $t^{'}$. 
$t^{'}$ is then injected into SPADE ResBlocks (SRES) combined in HumanDiffusion. 
Meanwhile, the semantic map $S$ is spatially disentangled into $seg$ and injected into SRES as the pose condition.
$F_{p1}$ is then transferred into a Self-Attentive Alignment module (SAA) to obtain the highlight features $F_{ps}$. Then $F_{ps}$ is sent into MCA. 
The MCA module performs Cross-modality Attentive Alignment (CAA) between $F_{ps}$ and the middle image features $f_{i}$ in SRES, from low-resolution to high-resolution. 
Finally, the refined human features are obtained for finer image generation.

For the sampling process, the desired pose map and the text description are treated as inputs. 
Then a satisfying human image could be obtained. 
In the following, we will describe the significant modules in detail.

\subsection{Stylized Memory Retrieval} 
\begin{figure}[t]
\centering
\includegraphics[width=0.48\textwidth]{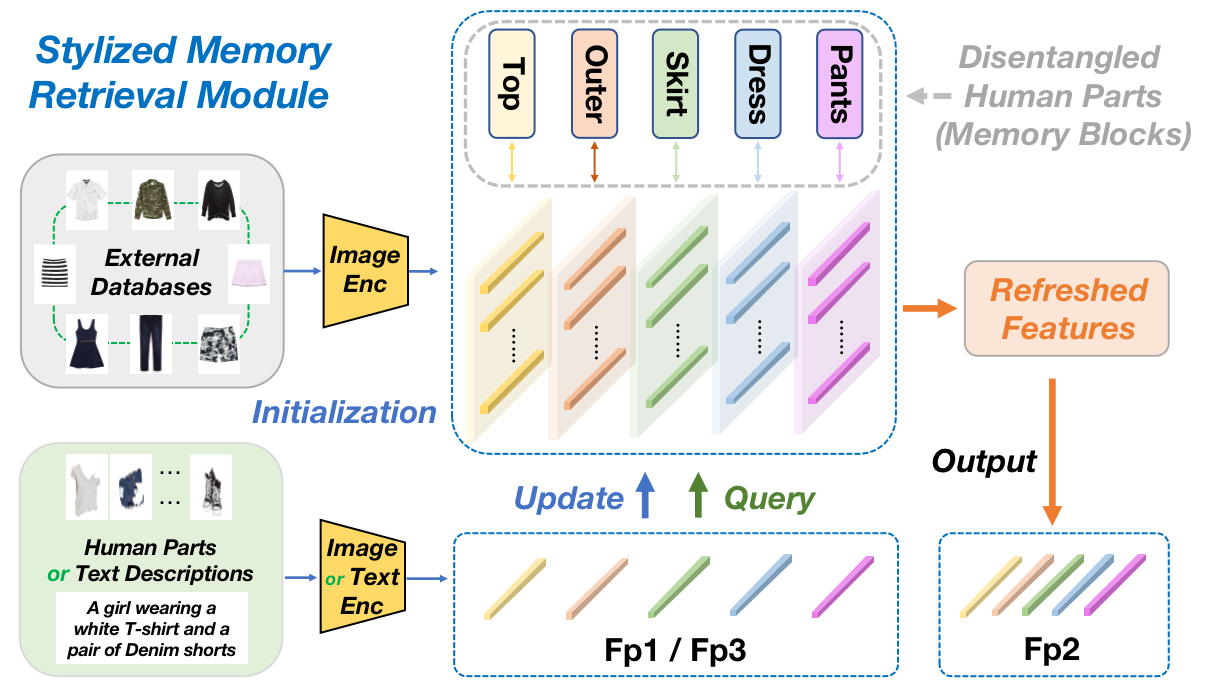} 
\caption{Stylized Memory Retrieval Module.}
\label{Figure3}
\vspace{-3mm}
\end{figure}
To generate high-quality person images, appearance features act as the most important prior.
To obtain fine-grained appearance features as the global condition of the diffusion, we propose the Stylized Memory Retrieval (SMR) module in data processing, as shown in Figure \ref{Figure3}.
We design disentangled memory blocks to store the image-aware CLIP features\cite{radford2021learning}, which corresponds to the specific human parts. 
There are two steps to establish this memory module, initialization with external databases and updating with the training dataset.
As shown in Figure \ref{Figure3}, the SMR is first initialized with the CLIP features of clothing images in free-style.
Then, the memory blocks are updated with the training data (\emph{i.e.} image CLIP features). 
With the external prototype clothing features and the specific characteristics from the training set, the SMR could provide fine-grained and image-aware representations of the input appearances.

\vspace{+1mm}
\noindent \textbf{Memory Initialization.} We establish the disentangled memory blocks for different semantic categories to store the semantic-aware features, where $n$ represents the number of disentangled human parts, and $c$ is the feature dimension. 
An external database\cite{liuLQWTcvpr16DeepFashion} is treated as the retrieval source, which contains clothes images in various styles. 500 images are selected as the memory slots for each category, which almost contain all clothing styles. 
Thus, the memory has been established with $n$ blocks ($M \in R^{n\times (m\times c)}$) and each block ($E \in R^{m\times c}$)  contains $m$ memory slots $e \in \mathbb{R}^c$.
The restored memory slots are encoded by the CLIP encoder.

\vspace{+1mm}
\noindent \textbf{Memory Updating.} 
In the training procedure, the memory blocks are fine-tuned with the DeepFashion training dataset gradually. 
Given a human image, the CLIP encoder maps the image into latent representations $F \in R^{n\times c}$ ($F_{p1}$), which represents the disentangled parts of this human. 
We use the similarity between the input latent feature and the relevant memory slots to update the memory blocks. 
For $ith$ semantic in the image, we compute the cosine distance $s_{ij}$, between $F_{i}$ and the $jth$ memory slot $e_{ij}$, which is formulated as follows:
\begin{equation}
s_{i j}=\frac{e_{i j} F_i^T}{\left\|e_{i j}\right\|\left\|F_i\right\|}
\label{equation1}
\end{equation}
Then, we seek the memory slot $e_{ik}$, which is most relevant with the input query $F_{i}$:
\begin{equation}
k=\underset{j}{\operatorname{argmax}}\left(s_{i j}\right)
\end{equation}
To update the relevance between the memory with the training images, we use the latent vector $F_{i}$ in each batch, to update the memory slot $e_{ik}$, which also represents the initial features for the query $F_{i}$. The memory updating process can be formulated as follows:
\begin{equation}
\hat{e_{i k}} \leftarrow \beta e_{i k}+(1-\beta) F_i
\end{equation}
where $\beta \in [0,1]$ is a decay rate set to 0.995 in our paper. Thus, the memory blocks integrate the prototype cloth features and the specific characteristics of the training set.\\
\vspace{-3mm}

\noindent \textbf{Memory Reading.}  
After the memory updating, we refresh the input latent vector $M_{i}$ to a fine-grained representation $\hat{M}_i$.
We employ an attentive aggregation training strategy. 
To begin with, the cosine similarity matrix $\gamma$ = $\left\{\gamma_{i j} \mid i=1, \ldots, n, j=1, \ldots, m\right\}$ is computed through Equation \ref{equation1} again. 
Then, the soft scores $A$=$\left\{a_{i j} \mid i=1, \ldots, n, j=1, \ldots, m\right\}$ are formulated by a softmax operation.
\begin{equation}
a_{i j}=\frac{\exp \left(\gamma_{i j}\right)}{\sum_{j=1}^m \exp \left(\gamma_{i j}\right)}
\end{equation}
Finally, the memory-based latent vectors $\hat{F}_i$ ($F_{p2}$) are constructed by aggregating memory slots with soft scores.
\begin{equation}
\hat{F}_i=\sum_j^m a_{i j} \hat{e_{i j}}
\end{equation}

To summarize, we propose a Stylized Memory Retrieval Module which utilizes an external database to get fine-grained image-aware feature vectors $F_{p2}$ which provides abundant clothing representations in free-style. 
The SMR module is  jointly trained with the diffusion networks in the following. 
Furthermore, while sampling, the retrieval operation helps to refresh the text CLIP features to be image-aware, which provides the potential for text-free training.

\subsection{HumanDiffusion: Training}
In this section, we will introduce the training procedure of the HumanDiffusion framework specifically.
\vspace{+1mm}

\noindent \textbf{Data Processing.} 
For each training iteration, we first perform the dot multiplication on the input image $I$ and the corresponding semantic map $S$ to $n$ parts $I_{i}$ (specific clothes appearances). 
Then, all the $I_{i}\in I$ is sent into the CLIP image encoder to extract a $n \times 512$ features:
\begin{equation}
I_{i} = I \otimes S_{i}
\end{equation}
\begin{equation}
F_{p1i} = CLIP(I_{i})
\end{equation}
where $S_{i}$ denotes one-part semantic map in human images, $F_{p1i}$ is the partial CLIP features in $F_{p1}$. 
Then the $F_{p1}$ is sent into the SMR module to get refreshed features $F_{p2}$ which are fused with the time step $t$ to act as the global condition $t^{'}$. The $seg$ from semantic map $S$ is acted as the pose condition and is also injected into the SPADE ResBlocks.

\begin{figure}[t]
\centering\includegraphics[width=0.42\textwidth]{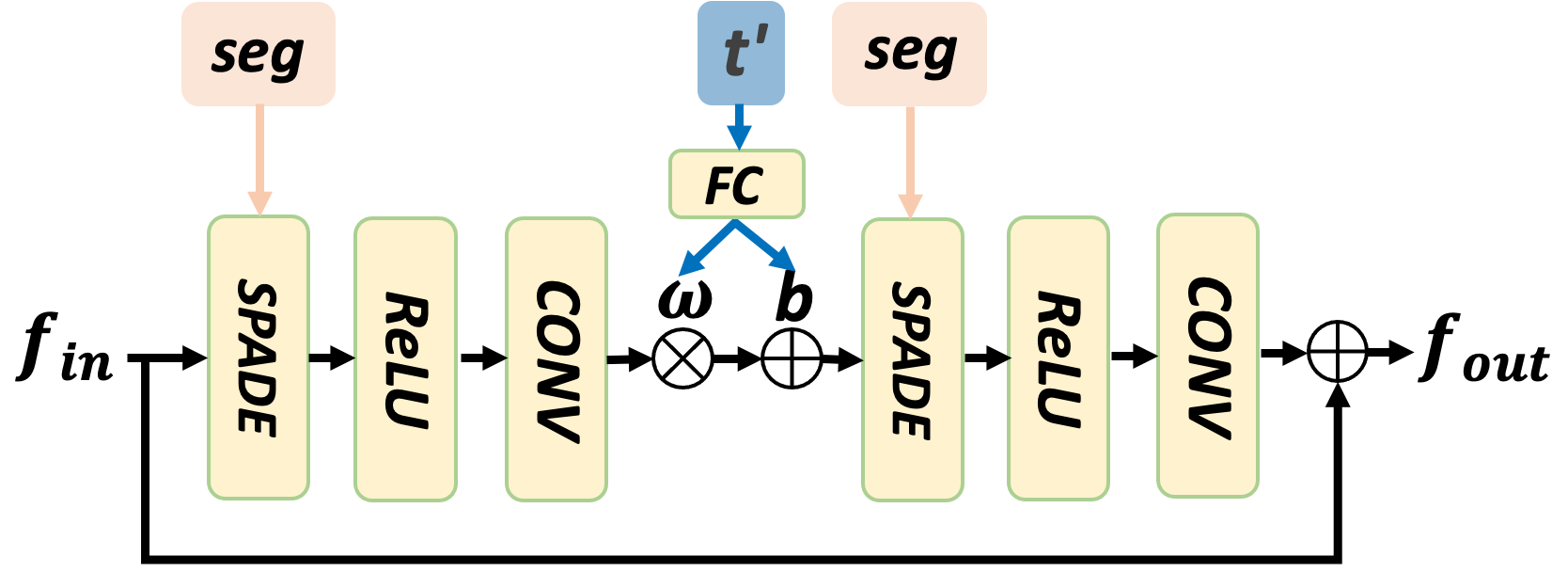} 
\caption{The architecture of SPADE ResBlock.}
\vspace{-3mm}
\label{Figure4}
\end{figure}

\vspace{+1mm}
\noindent \textbf{SPADE ResBlocks.}
Inspired by \cite{wang2022semantic,park2019semantic}, we design the SPADE ResBlocks (SRES) to inject the pose semantic information $seg$ and the global condition $t^{'}$ for human pose control and image-aware feature guidance. 
As shown in Figure \ref{Figure4}, the SRES module employs spatially adaptive normalization (SPADE)\cite{park2019semantic} to inject the pose feature $seg$.

\begin{equation}
f_{spade}=SPADE(f_{in}, seg)
\label{8}
\end{equation}
\begin{equation}
f_{t^{'}}=CONV(ReLU(f_{spade})) \times (1+\omega) + b
\label{9}
\end{equation}
As shown in Equation \ref{8}, $f_{in}$ is the input features of SRES, and $f_{spade}$ is the output of SPADE which contains more pose information. 
$t^{'}$ is injected as shown in Equation \ref{9}, in which $\omega$ and $b$ are modulation coefficients
obtained by sending the global condition $t^{'}$ into Fully Connected (FC) layers. 
Then $f_{t^{'}}$ contains more fine-grained appearance features.
In summary, the SRES module helps the diffusion model to obtain more pose and image-aware feature information.
\begin{figure}[t]
\centering
\includegraphics[width=0.45\textwidth]{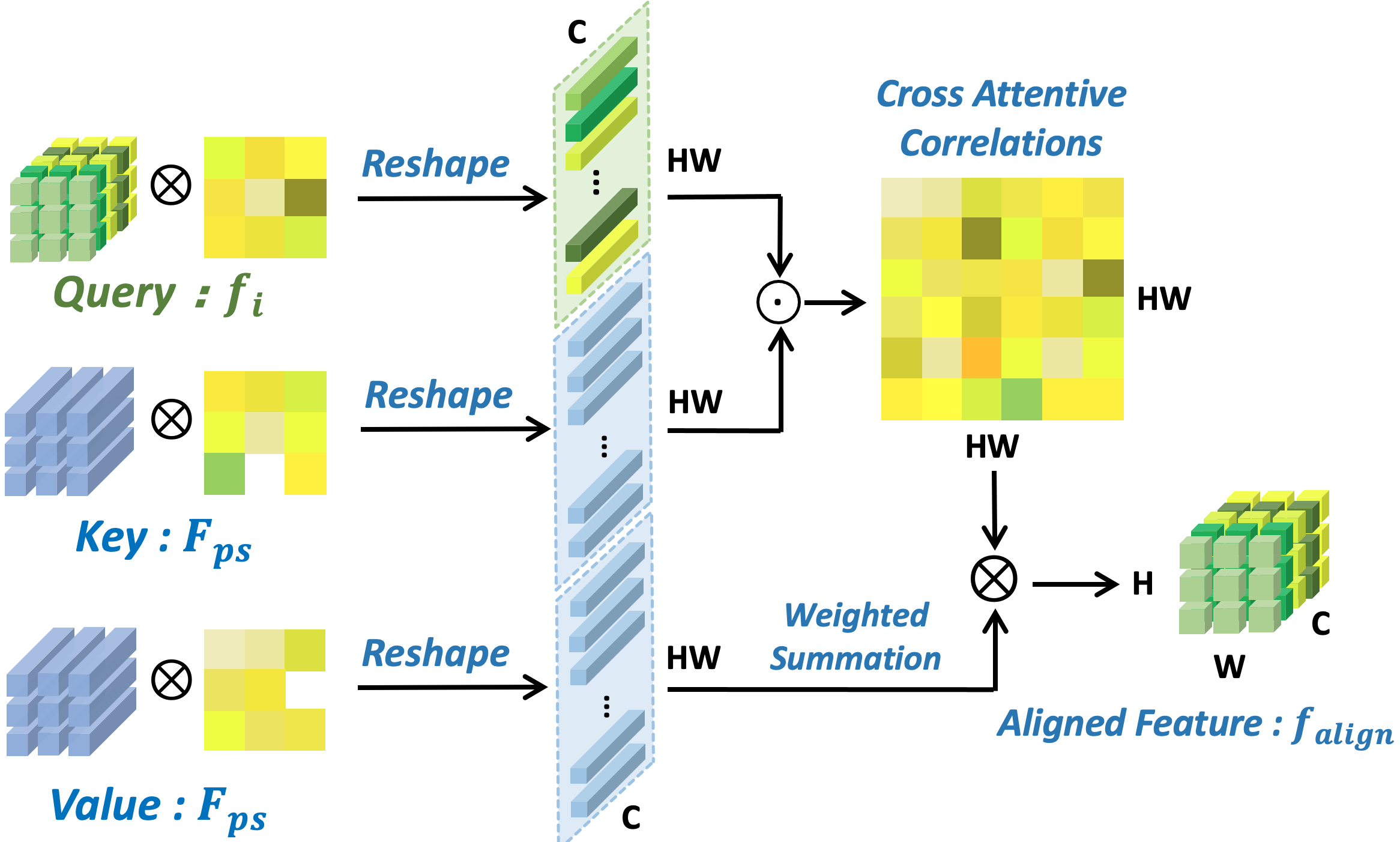} 
\caption{Cross-modality Attentive Alignment Module.}
\vspace{-4mm}
\label{Figure5}
\end{figure}

\noindent \textbf{Multi-scale Cross-modality Alignment.}  Previous general text-to-image researches like Dalle2\cite{ramesh2022hierarchical} and GLIDE\cite{nichol2021glide} mostly utilize the powerful coding capabilities of the CLIP to measure and align the distance between the output images and the input prompts. 
For human image generation, human images have strong structural information. 
It is difficult to generate image details of complex structures depending on simple alignment strategies or constraints.
As shown in Figure \ref{Figure5}, we design a Cross-modality Attentive Alignment (CAA) module which constitutes MCA in the multi-scale resolutions of the diffusion backbone, to gradually align the fine-grained partial features $F_{ps}$ with the U-Net middle features $f_{i}$. In this module, we treat $F_{p1}$ as the text-aware encoding and $F_{p2}$ as the image-aware encoding.

Before the MCA, for distilling finer features that highlight the important parts of $F_{p1}$, we perform self-attention to obtain fine-grained partial features $F_{ps}$.
Then, the MCA module will further align the cross-modality features from coarse to fine.
CAA is carried out through multi-scale resolutions from 8 $\times$ 8 to 32 $\times$ 32.
\begin{figure*}[t]
\centering
\includegraphics[width=1.005\textwidth]{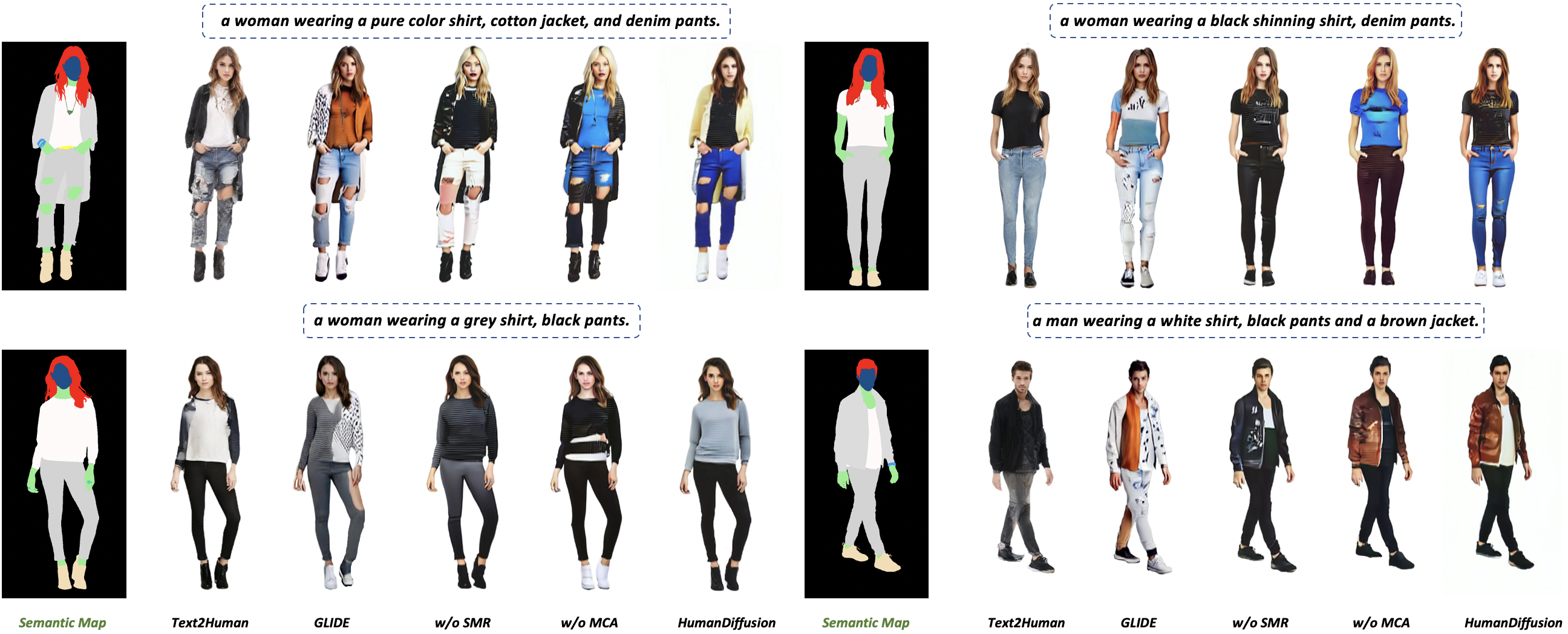} 
\caption{Quality Comparisons on image generation under the guidance of text on Text2Human, GLIDE, HumanDiffusion, and the ablations of HumanDiffusion without SMR and MCA. HumanDiffusion could express human appearance in open-vocabulary and free-style words. Compared to other methods, HumanDiffusion's generation results are mostly corresponding to descriptions. \textbf{Zoom in} for better details.}
\vspace{-4mm}
\label{Figure6}
\end{figure*}
Specifically, as shown in Figure \ref{Figure5}, CAA has three learnable projection matrices $W_{Q}$, $W_{K}$, $W_{V}$. The middle features $f_{i}$ are treated as queries $Q\in \mathbb{R}^{b \times r \times r}$, where $b$ is the batchsize, $r$ is the resolution of middle features. 
The fine-grained features $F_{ps}$ from SAA are performed as the keys $K$ and values $V$, which are queried by $Q$ for alignment information distillation:
\begin{equation}
    Q_{i} =f_{i} W_Q, K=F_{ps} W_K, V=F_{ps} W_V
\label{10}
\end{equation}
\begin{equation}
\begin{aligned}
    f_{align} & =CAA(Q_{i}, K, V)\\
&              =Softmax\left(Q_{i} K^{\mathrm{T}} / \sqrt{C}\right) V 
\end{aligned}
\label{11}
\end{equation}
\begin{equation}
    f_{i}^{'} = f_{i} + f_{align}
\label{12}
\end{equation}
Note $f_{align}$ is the aligned features. $f_{i}^{'}$ is the output features that are sent into the next SRES.

After the coarse-to-fine alignment operations in Equations \ref{10}-\ref{12}, the diffusion is more sensitive to spatial information, and each part of the features could obtain better correspondences with its text descriptions. 
As a result, the denoising diffusion procedure could be improved as Equation \ref{13}.
In each step at time step = $t$, $x_{t-1}$ and $x_{t}$ are the input and output image maps, $t^{'}$ and $seg$ are the global and pose conditions, $Fp$ is the text-aware features, $\mu_\theta$ and $\Sigma_\theta$ are the mean and variance of the output which is decided by the U-Net parameter $\theta$. 
\vspace{-4mm}

\begin{equation}
\centering
\begin{aligned}
p_\theta\left(x_{t-1} \mid x_t\right)
&= \mathcal{N}(x_{t-1} ; \mu_\theta(x_t, (t^{'}, seg, Fp)), \\
&\Sigma_\theta(x_t, (t^{'}, seg, Fp))
\end{aligned}
\label{13}
\end{equation}

\subsection{HumanDiffusion: Sampling}
\vspace{-1mm}
As shown in the bottom of Figure \ref{Figure2}, for human image sampling, the target semantic map and the desired open-vocabulary text description is sent into the framework. 
First, the text description is disentangled according to the partial image structure and then encoded by the CLIP text encoder to get partial text features $F_{p3}$. Then SMR refreshes $F_{p3}$ and gets the fine-grained image-aware information, which promotes HumanDiffusion to use only images for training and open-vocabulary texts for sampling.

Similar to the training process, the initial features ($F_{p3}$), refined text features, and the semantic pose map are all inputted into HumanDiffusion to generate the target person.
Finally, a human image that conforms to the text description and appears in the target pose is obtained.
\vspace{-2mm}
\section{Experiments}
\noindent \textbf{Dataset.}  We conduct our experiments on DeepFashion-MultiModal dataset\cite{liuLQWTcvpr16DeepFashion} to verify the effectiveness of our method. As for the external memory database, we employ the In-shop Clothes Retrieval database\cite{liuLQWTcvpr16DeepFashion}, which contains large amounts of partial clothes images with clear detailed photographs. 
The DeepFashion MultiModal dataset contains 12701 full-body images with the corresponding semantic maps and text descriptions.
The semantic maps contain parsing labels of 24 classes. 
We follow the setup of Text2Human\cite{jiang2022text2human} to spilt the training and testing set.

\noindent \textbf{Implementation Details.} 
The diffusion model is implemented by PyTorch \cite{li2020pytorch} with 4 NVIDIA Tesla-A6000 GPUs under the batchsize 24. 
We use Adam optimizer \cite{kingma2014adam} as the gradient descent method, and set the learning rate $\alpha=0.0001$. 
The model is trained for 240000 iterations and finetuned after 100000 iterations with the $\alpha$ set to 0.00001. We select 10 parts from the semantic maps, which can describe the basic human clothes appearances (background, top, outer, skirt, dress, pants, hair, face, shoes, skin). 
Similar to previous diffusion-based methods, we first train the model under 128$\times$256 resolution and then train a super-resolution upsampling model up to 256$\times$512. 

\noindent \textbf{Text-free Training.} HumanDiffusion can only use human images and semantic pose maps as inputs to train the model. With the help of SMR, whether the CLIP image features or text features ($F_{p1}/F_{p3}$) could be treated as text-aware embedding, whenever training or sampling.

\noindent \textbf{Evaluation Metrics.} Three objective indicators and a user study are used to evaluate the generation performance. Frechet Inception
Distance (FID) measures the fidelity of the generated images \cite{heusel2017gans}. Learned Perceptual Image Patch Similarity (LPIPS) computes the distance between the generated images and ground truths in the perceptual domain \cite{zhang2018unreasonable}. Multi-Scale Structural Similarity (MS-SSIM) evaluates the structural similarity between two images \cite{wang2003multiscale}. 
The user study is a subjective evaluation of the image quality and caption similarity.
\begin{figure*}[t]
\centering
\includegraphics[width=0.96\textwidth]{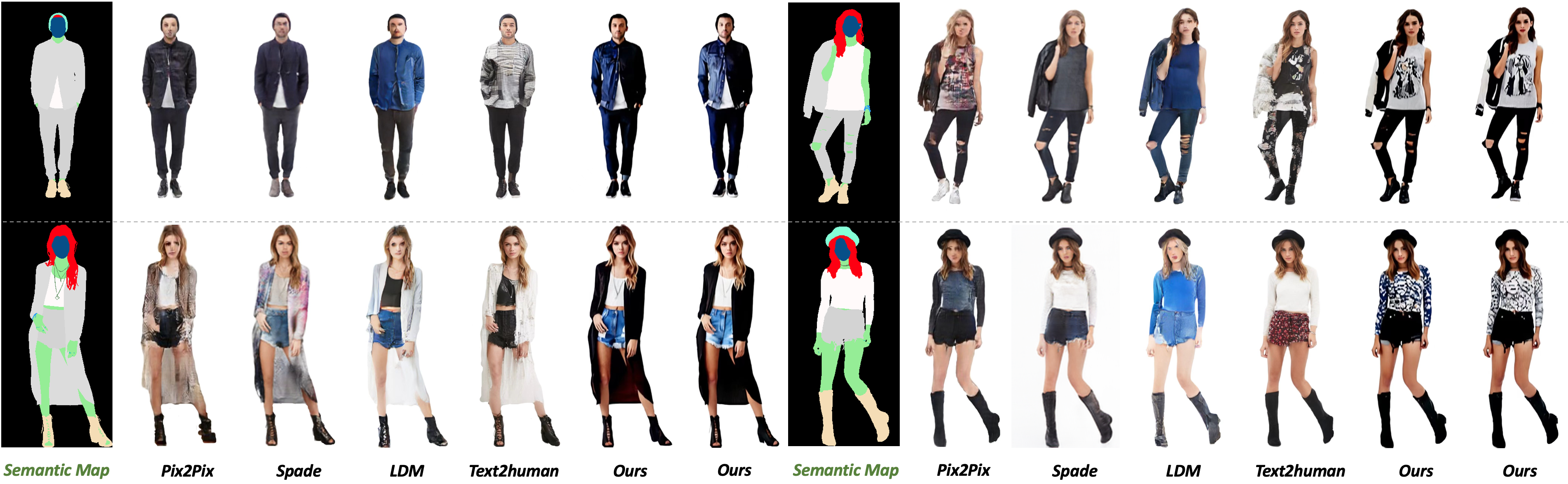} 
\caption{Quality Comparisons on image generation under the guidance of semantic. The leftmost of each group is the input semantic map, columns from left to right represent the results of Pix2Pix, Spade, LatentDiffusion, Text2Human, and two sampled images by HumanDiffusion. HumanDiffusion can generate diverse human images in high-quality. \textbf{Zoom in} for better details. }
\label{Figure7}
\vspace{-3mm}
\end{figure*}
\vspace{-2mm}
\subsection{Text-Driven Person Image Generation}
\subsubsection{Text-Driven Quantitative Results}
\vspace{-2mm} 
As shown in Table \ref{tab1}, HumanDiffusion obtains competitive results with Text2Human in FID and outperforms it in LPIPS and MS-SSIM, where HumanDiffusion employs a more compact and flexible framework. Furthermore, the results from GLIDE and ablation experiments show that the SMR and MCA modules could improve human images in terms of image quality and structural similarity. 
\begin{table}[t]
    \centering
    \caption{Quantitative comparisons and ablations on Text-Driven Person Image Generation with DeepFashion dataset.}
    \begin{tabular}{cl cc cc cc }
    \hline
         Methods & FID$\downarrow$ & LPIPS$\downarrow$ & MS-SSIM$\uparrow$ \\
    \hline
          Text2Human  &  \textbf{29.53}  & 0.2154 & 0.7358        \\
          GLIDE & 56.53 & 0.3215 & 0.5658 \\
          w/o SMR  &  38.43  &  0.2224 & 0.7326       \\
          w/o MCA  &  34.86  &  0.2382 & 0.7022       \\ 
    \hline
          \rowcolor{gray!20}
         \textbf{HumanDiffusion} &  30.42  &  \textbf{0.2095} & \textbf{0.7453}\\
    \hline
\end{tabular}
\label{tab1}
\vspace{-5mm}
\end{table}


\subsubsection{Text-Driven Qualitative Results}
\vspace{-2mm}
As shown in Figure \ref{Figure1} and \ref{Figure6}, given a target pose and the appearance text description, the generation results are obtained for illustration. 
It is worth noting that Text2Human uses only preset words (\emph{e.g.} pure color, denim, floral) to describe human appearances. 
HumanDiffusion could use free-style words ($e.g.$ `shinning', `colorful stripes', `red floral') to describe appearances, which can generate corresponding images in high-quality, as shown in Figure \ref{Figure1}.
For a fair comparison, we employ similar descriptions in Figure \ref{Figure6}, which Text2Human uses to generate. 
HumanDiffusion could obtain comparable results with a better match with the text descriptions.
Meanwhile, HumanDiffusion also performs better details and clearer disentanglement in different regions ($e.g.$ generating high-fidelity hands, faces, and clear semantic boundaries), whereas Text2Human occurs distortions or blends in different human parts (e.g. the mix of the shirt/pants or the hands/pants).
We also conduct experiments on the general text-to-image diffusion model GLIDE, which is hard to generate fine-grained features when the text expression is complicated. 
In addition, ablation experiments are also conducted.
As shown in Figure \ref{Figure6}, the models without SMR and MCA can not represent the appearance style (e.g. denim) correctly and align the image features with the text descriptions suitably (e.g. the shirt partly appearing denim and the pants partly appearing black).
In summary, HumanDiffusion can use open-vocabulary descriptions with semantic poses to flexibly generate high-quality human images in free-style.
\vspace{-1mm}
\begin{figure}[H]
\centering
\includegraphics[width=0.48\textwidth]{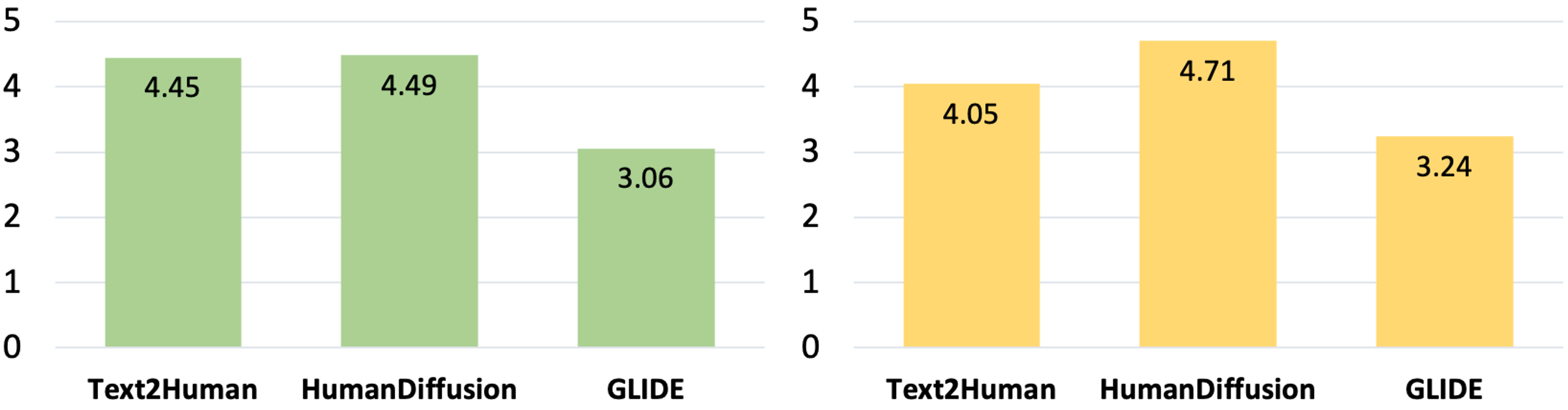}
\caption{User Study Results. (Left) Scores of photorealism of generated images. (Right) Scores of the similarity between the caption and the generation image. }
\vspace{-5mm}
\label{Figure8}
\end{figure}

\subsubsection{User Study}
\vspace{-2mm}
As Figure \ref{Figure8} shows, the User Study evaluation is performed to evaluate the quality of the generated images. 
50 users are presented with 20 groups of results, in which each group has three images generated by Text2Human GLIDE and HumanDiffusion.
They asked to rank 1) image photorealism and 2) caption similarity of the generated images (The scores rank is 3/4/5, higher is better). As a result, HumanDiffusion obtains higher scores for both the photorealism and captions similarity.
\vspace{-1mm}

\subsection{Semantic-Driven Person Image Generation} 
\subsubsection{Semantic-Driven Quantitative Results}
\vspace{-1mm}
As shown in Table \ref{tab2}, HumanDiffusion achieves satisfactory scores in the three indicators. 
HumanDiffusion outperforms Pix2Pix\cite{wang2018high}, SPADE\cite{park2019semantic} and LatentDiffusion (LDM)\cite{rombach2022high}, and exceeds the Text2Human with an improvement of on LPIPS and MSSSIM. 
The lightweight of our model limits the quality of generation to some extent, which greatly simplifies the training procedure.
\vspace{-1mm}

\begin{table}[H]
    \centering
    \caption{Quantitative comparisons on Semantic-Driven Person Image Generation with DeepFashion dataset.}
    \begin{tabular}{cl cc cc cc }
    \hline
         Methods & FID$\downarrow$ & LPIPS$\downarrow$ & MS-SSIM$\uparrow$ \\
    \hline
          Pix2Pix  &  38.46  & 0.2164 & 0.7293        \\
          SPADE  &  40.53  &  0.2327 & 0.7668       \\
          Latent Diffusion & 36.58 & 0.2143 & 0.7436\\
          Text2Human  & \textbf{30.89}  &  0.2045 & 0.7414   \\ 
    \hline
          \rowcolor{gray!20}
         \textbf{HumanDiffusion}  &  31.28  &  \textbf{0.2001} & \textbf{0.7736}\\ 
         \hline
\end{tabular}
\label{tab2}
\vspace{-2mm}
\end{table}

\subsubsection{Semantic-Driven Qualitative Results}
\vspace{-2mm}
As Figure \ref{Figure7} shows, Pix2Pix, SPADE, and LDM can synthesize person images in the target pose, but in low quality with distortion in details. 
Text2Human and HumanDiffusion generate images in high quality, while HumanDiffusion can generate diverse images (e.g. a person with several similar but different appearances in the same pose).
HumanDiffusion also performs better in details like the edge of clothes, and the distinction between clothing parts and body parts, which proves that HumanDiffusion performs satisfactorily in image quality and structural rationality.
\begin{figure}
\centering
\includegraphics[width=0.483\textwidth]{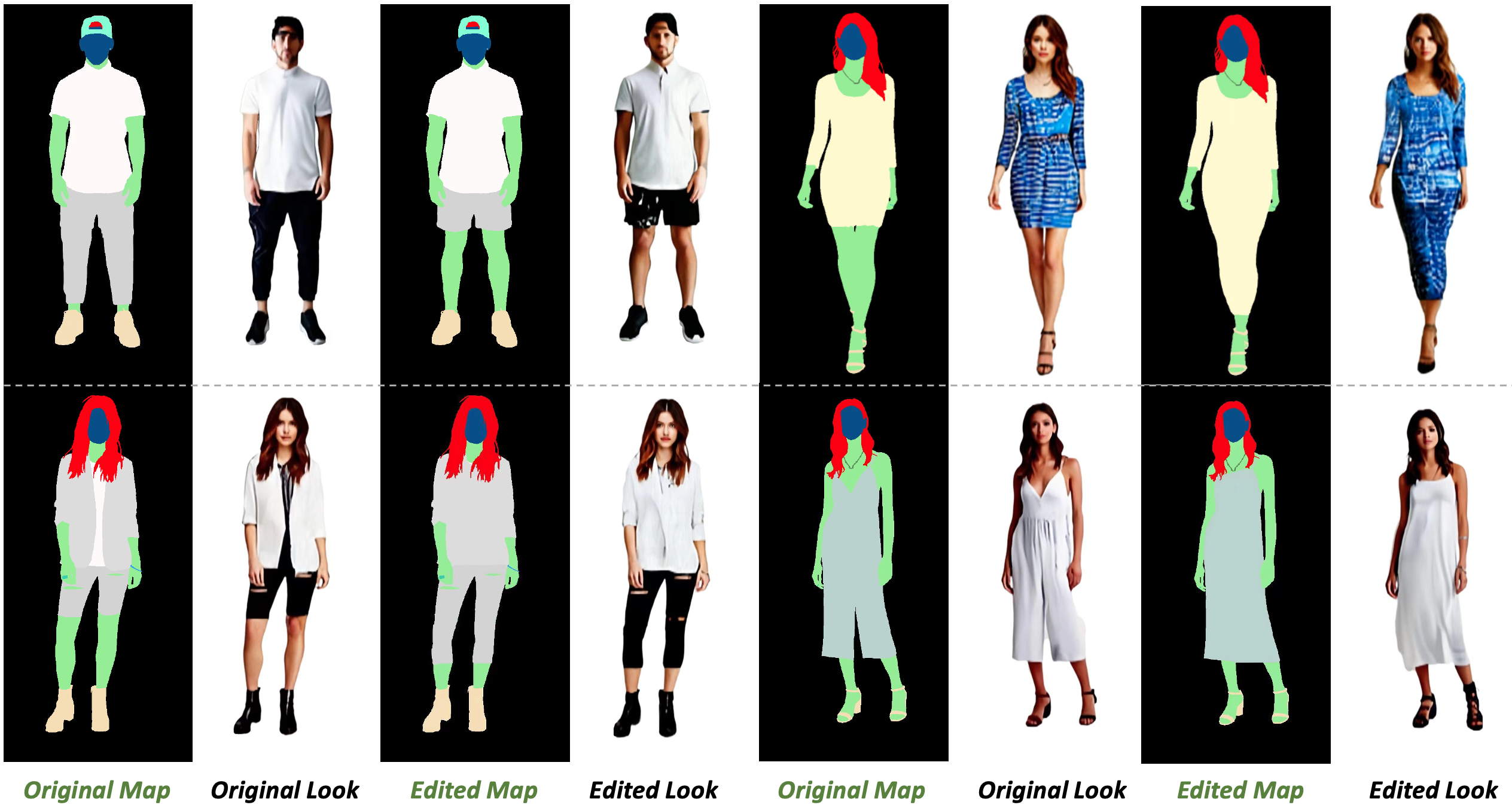}
\caption{Semantic-driven human partially editing.}
\label{Figure9}
\end{figure}
\vspace{-4mm}
\begin{figure}
\centering
\includegraphics[width=0.5\textwidth]{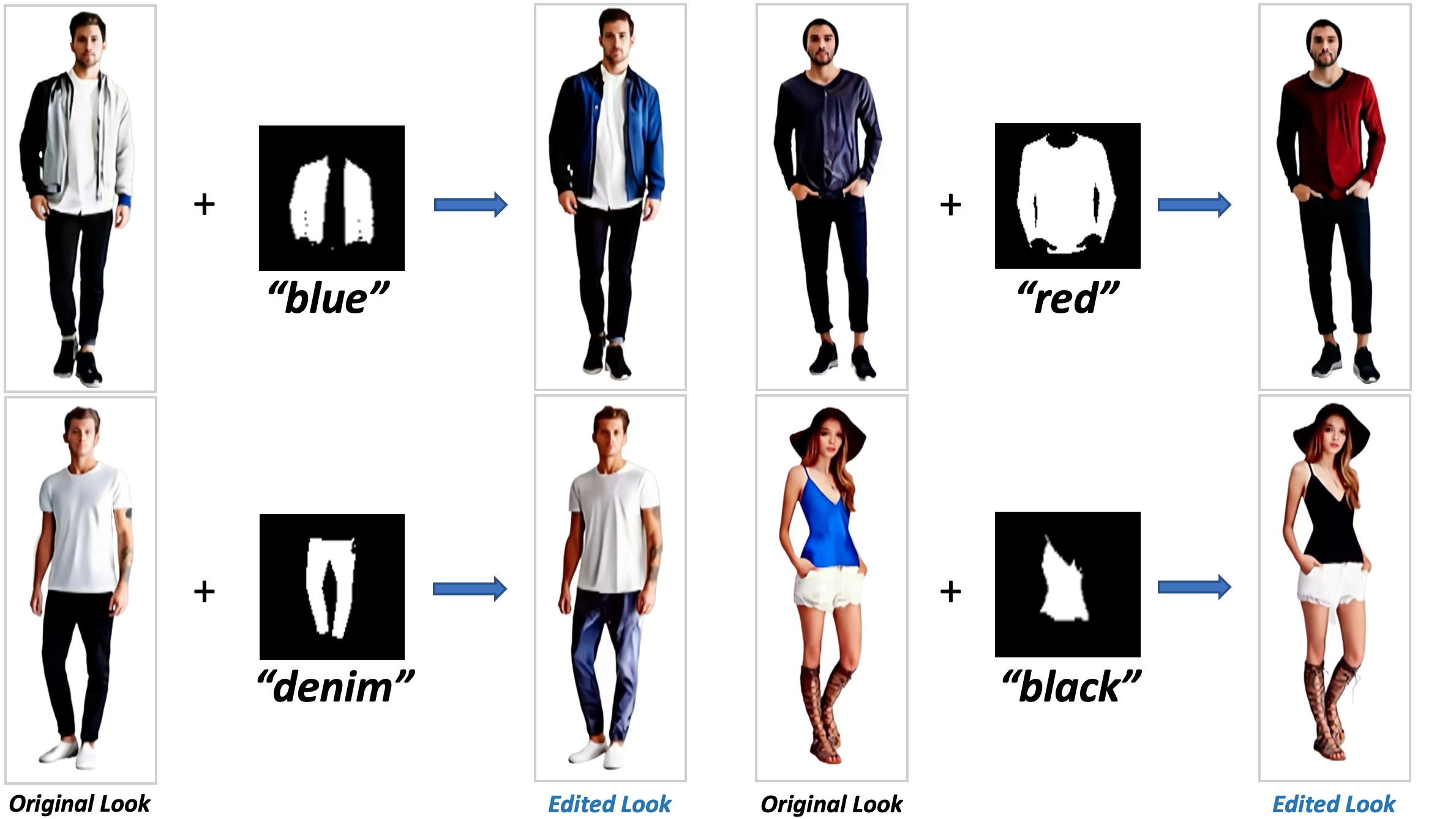}
\caption{Text-driven human partially editing.}
\vspace{-2mm}
\label{Figure10}
\end{figure}
\vspace{+4mm}
\subsection{Human Partially Editing}
In this section, we will introduce the performance of our method in editable applications. 
Different from previous generation models, we propose a more flexible method that allows using \textbf{text} or \textbf{semantic} to edit human appearances. 
As shown in Figure \ref{Figure9}, we can change a man in long pants to a man in shorts by editing the pants semantic map. Slight edit on a woman in rompers can change her to wear a dress in the same style and same pose. 
Our method can edit local areas while preserving the rest regions unchanged.

Meanwhile, we can also edit human images under text guidance. For example, a man's jacket color is aimed to change from grey to blue. Inspired by Blended-Diffusion\cite{avrahami2022blended}, we perform a \textbf{global denoising} process and a \textbf{local denoising} process simultaneously which aims to decrease the distance between the prompts (`blue') and the mask image (jacket) features in CLIP distance. 
Stitching operations are performed on them which helps to edit the mask region and preserve the rest unchanged. 
As shown in Figure \ref{Figure10}, we can generate edited images conforming to the text descriptions.
\begin{figure}
\centering
\includegraphics[width=0.418\textwidth]{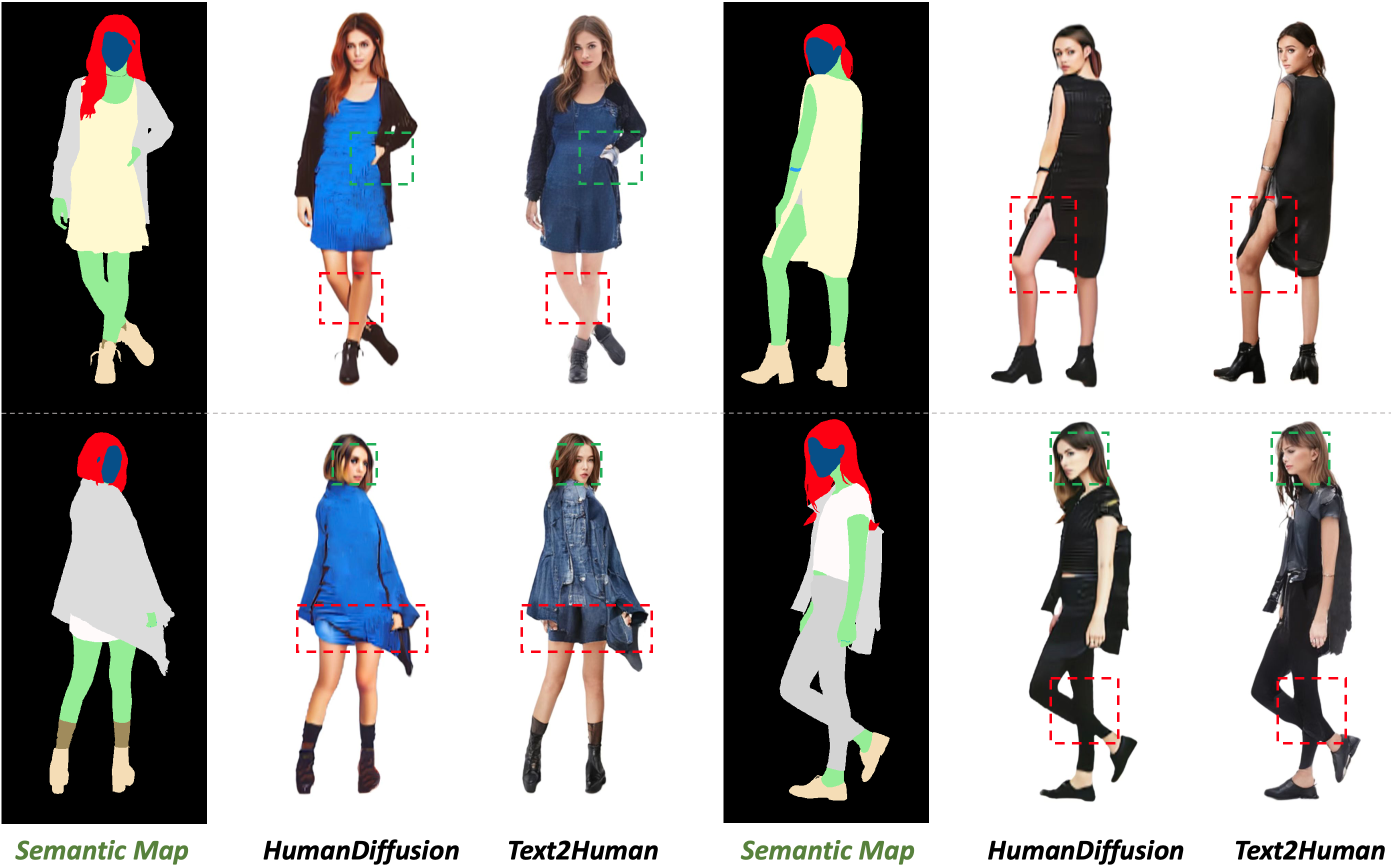}
\caption{Comparisons on Text-driven complicated person generation. Please \textbf{zoom in} for details.}
\vspace{-2mm}
\label{Figure11}
\end{figure}
\subsection{Complicated Person Generation}
\vspace{-1mm}
Existing human generation methods mostly perform poorly in \textbf{complicated person generation} (\emph{e.g.} Occlusion situations: the person showing a side face or standing with legs crosses). 
As shown in Figure \ref{Figure11}, HumanDiffusion shows better results compared to Text2Human. 
As shown in red dotted boxes, in the cross-leg example, Text2Human mixes the two legs.  
The side face generation and hands details shown in green dotted boxes also perform unnaturally distortion. 
In contrast, HumanDiffusion can identify and generate partial crossover areas and natural side faces.

\section{Limitations}
\vspace{-2mm}
There are two main limitations of our method.
First, there are blurred textures on some little partial parts (e.g. face, shoes), since our method concentrates more on body appearances and the SMR does not set specific memory blocks for all regions.
Second, some images have a cartoon style, which can be treated as a kind of blur.
This is probably caused by the inherent training of diffusion.
Please see more details in \textbf{S}upplementary \textbf{M}aterials. 

\section{Conclusion}
\vspace{-2mm}
In this paper, we propose a diffusion-based text-driven person image generation framework called HumanDiffusion. 
We design two collaborative modules, the Multi-scale Cross-modality Alignment module and the Stylized Memory Retrieval module to align and refine the image features from coarse to fine. Humandiffusion employs a compact architecture with end-to-end training, which could use open-vocabulary descriptions to generate person images in free-style and provide a more flexible and controllable way for real-world applications.

\clearpage
{\small
\bibliographystyle{ieee_fullname}
\bibliography{egbib}
}

\end{document}